\documentclass{article}

\usepackage[utf8]{inputenc}
\usepackage[preprint]{neurips_2025}
\usepackage{graphicx}
\usepackage{multirow} 
\usepackage{capt-of}
\usepackage{float}
\usepackage{array}
\usepackage{booktabs}
\usepackage{enumitem} 
\usepackage[table]{xcolor}
\usepackage{colortbl}
\usepackage{tikz}
\usepackage{adjustbox}
\usepackage[most]{tcolorbox}
\usepackage{subcaption} 
\usepackage{tabularx}
\usepackage{pifont}

\newcolumntype{C}[1]{>{\centering\arraybackslash}m{#1}}

\usepackage[utf8]{inputenc} % allow utf-8 input
\usepackage[T1]{fontenc}    % use 8-bit T1 fonts
\usepackage{hyperref}       % hyperlinks
\usepackage{url}            % simple URL typesetting
\usepackage{booktabs}       % professional-quality tables
\usepackage{amsfonts}       % blackboard math symbols
\usepackage{nicefrac}       % compact symbols for 1/2, etc.
\usepackage{microtype}      % microtypography
\usepackage{xcolor}         % colors
\usepackage{authblk}

\title{TerraIncognita: A Dynamic Benchmark for Species Discovery Using Frontier Models} 

\author[1]{Shivani Chiranjeevi}
\author[1]{Hossein Zaremehrjerdi}
\author[2]{Zi K. Deng}
\author[1]{Talukder Z. Jubery}
\author[3]{Ari Grele}
\author[1]{Arti Singh}
\author[1]{Asheesh K Singh}
\author[1]{Soumik Sarkar}
\author[2]{Nirav Merchant}
\author[4]{Harold F. Greeney}
\author[1]{Baskar Ganapathysubramanian}
\author[5]{Chinmay Hegde}

\affil[1]{Iowa State University, Ames, IA 50011, USA}
\affil[2]{University of Arizona, Tucson, AZ 85721, USA}
\affil[3]{University of Nevada, Reno, NV 89557, USA}
\affil[4]{Yanayacu Biological Station and Center for Creative Studies, Ecuador}
\affil[5]{New York University, New York, NY 10003, USA}
\affil[ ]{\texttt{chinmay.h@nyu.edu,baskarg@iastate.edu,antpittanest@gmail.com}}

\begin{document}

\maketitle

\begin{abstract} 
The rapid global loss of biodiversity, particularly among insects, represents an urgent ecological crisis. Current methods for insect species discovery are manual, slow, and severely constrained by taxonomic expertise, hindering timely conservation actions. We introduce TerraIncognita, a dynamic benchmark designed to evaluate state-of-the-art multimodal models for the challenging problem of identifying unknown, potentially undescribed insect species from image data. Our benchmark dataset combines a mix of expertly annotated images of insect species likely known to frontier AI models, and images of rare and poorly known species, for which few/no publicly available images exist. These images were collected from underexplored biodiversity hotspots, realistically mimicking open-world discovery scenarios faced by ecologists. The benchmark assesses models' proficiency in hierarchical taxonomic classification, their capability to detect and abstain from out-of-distribution (OOD) samples representing novel species, and their ability to generate explanations aligned with expert taxonomic knowledge. Notably, top-performing models achieve over 90\% F1 at the Order level on known species, but drop below 2\% at the Species level—highlighting the sharp difficulty gradient from coarse to fine taxonomic prediction (Order $\rightarrow$ Family $\rightarrow$ Genus $\rightarrow$ Species). Discovery accuracy on novel species varies widely, from 55\% to 88\%, revealing inconsistencies in abstention and overcommitment strategies across model families. Our evaluation of contemporary frontier multimodal models highlights critical strengths and exposes limitations in their current open-world recognition capabilities. TerraIncognita will be updated regularly, and by committing to quarterly dataset expansions (of both known and novel species), will provide an evolving platform for longitudinal benchmarking of frontier AI methods. All TerraIncognita data, results, and future updates are available \href{https://baskargroup.github.io/TerraIncognita/}{here}.
\end{abstract}
%NOTE: The figures currently in the draft are placeholders to illustrate the structure and will be updated with real numbers once the data is finalized. After sorting, I’ll run the models on the processed dataset and include the results. As of now, I have access to all four models mentioned in the paper.
\begin{figure}[hbtp]
    \centering
    \includegraphics[width=0.85\textwidth]{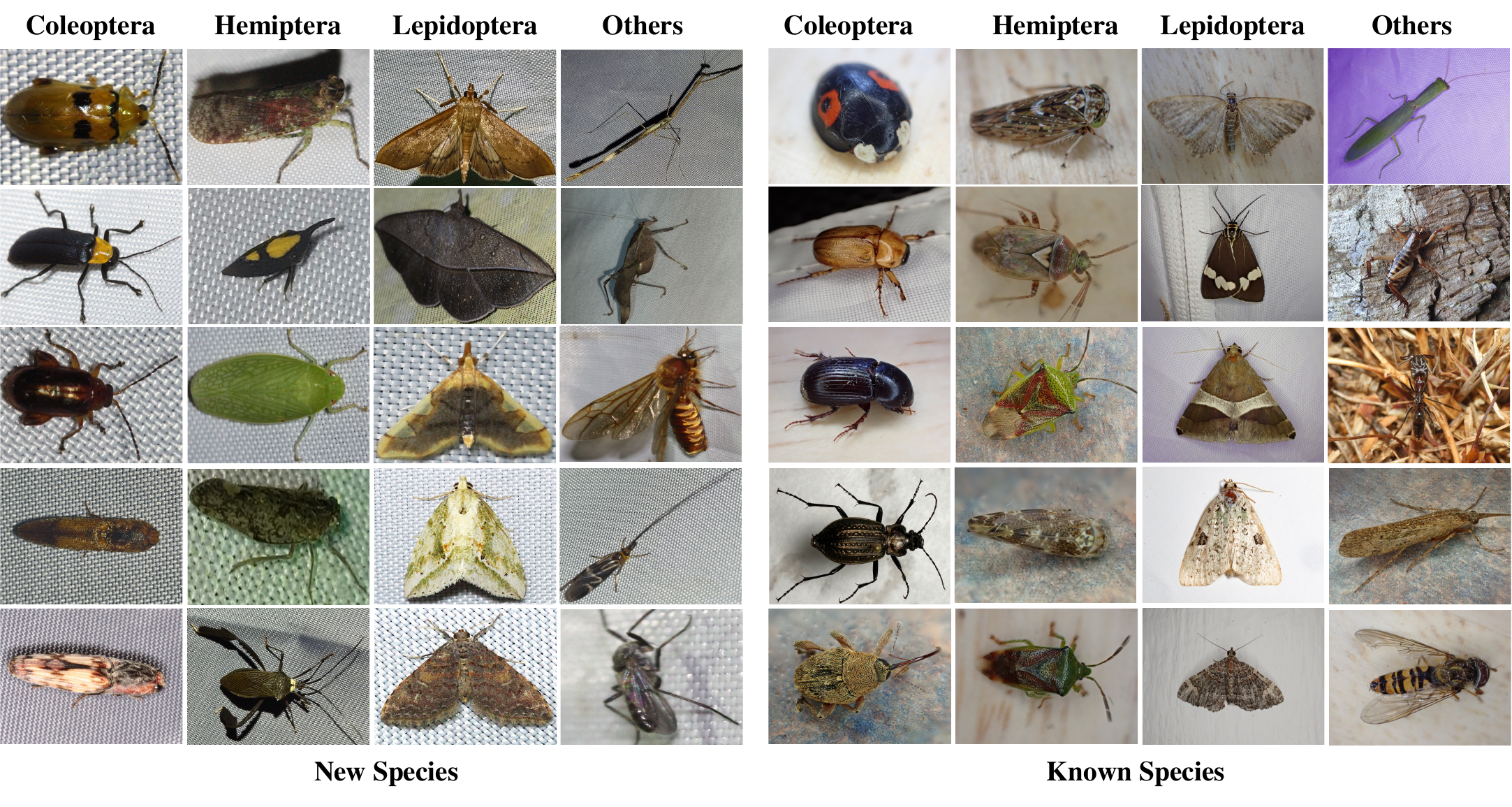}
    \caption{A few example images from the \textbf{TerraIncognita} dataset. The dataset comprises both \textit{known} as well as \textit{novel} insect species. Each column represents a taxonomic order; each row is an image, with the columns on the left showing \textit{Novel} species and those on the right showing \textit{Known} ones.}
    \label{fig:vlm-grid-orders}
\end{figure}

\begin{figure}[!t]
    \centering
    \includegraphics[height=0.725\textheight,keepaspectratio]{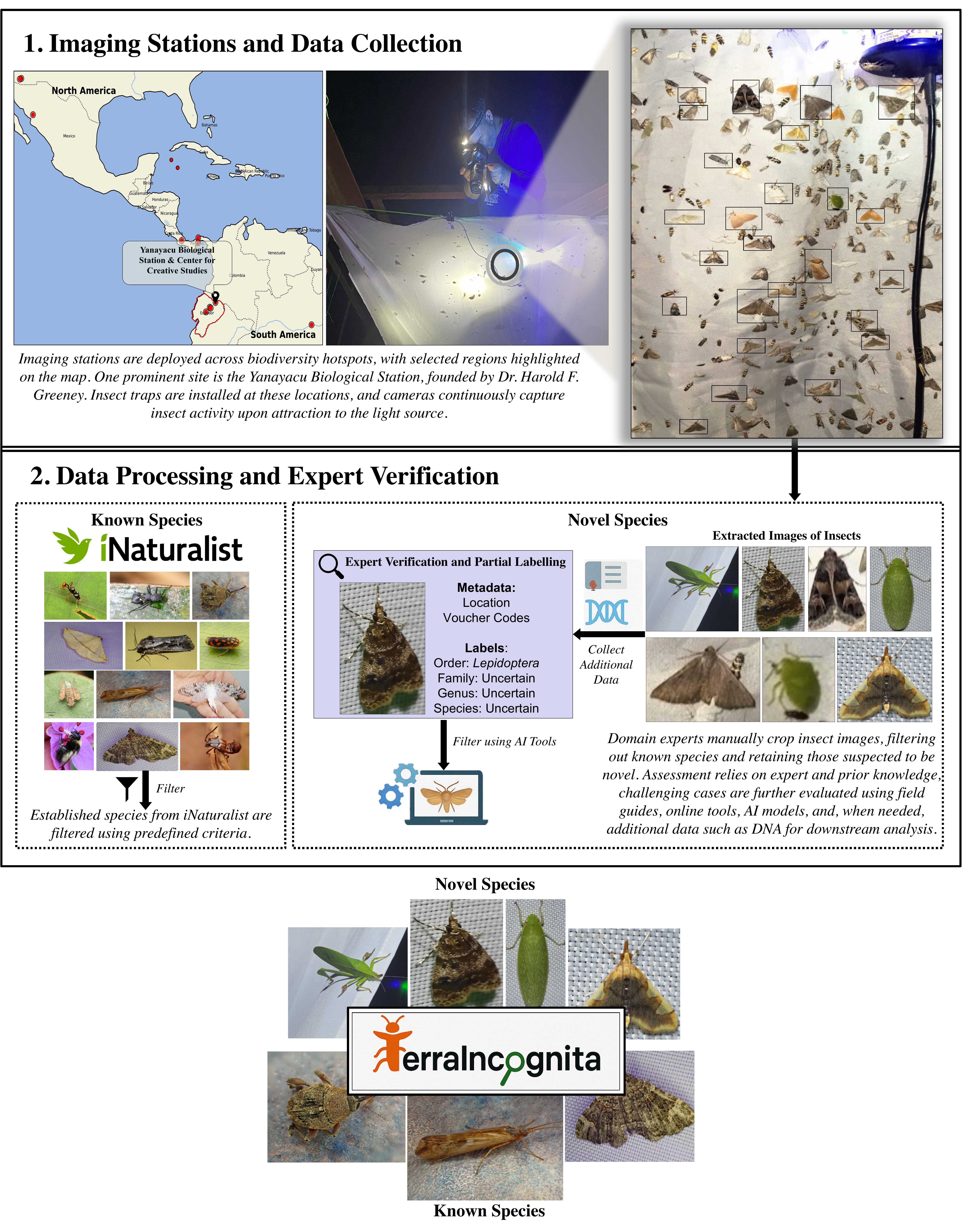}
    \caption{Overview of the TerraIncognita data collection pipeline. Insect specimens are first captured using light traps and examined by an entomologist or field expert. Images are then filtered using commercial-grade image classification tools to detect known taxonomic matches. If a specimen does not match existing species-level entries, it is flagged as potentially novel.}
    \label{fig:fullpage}
\end{figure}

\begin{figure}[t]
    \centering
    \includegraphics[width=0.7\textwidth, trim=15 15 0 0, clip]{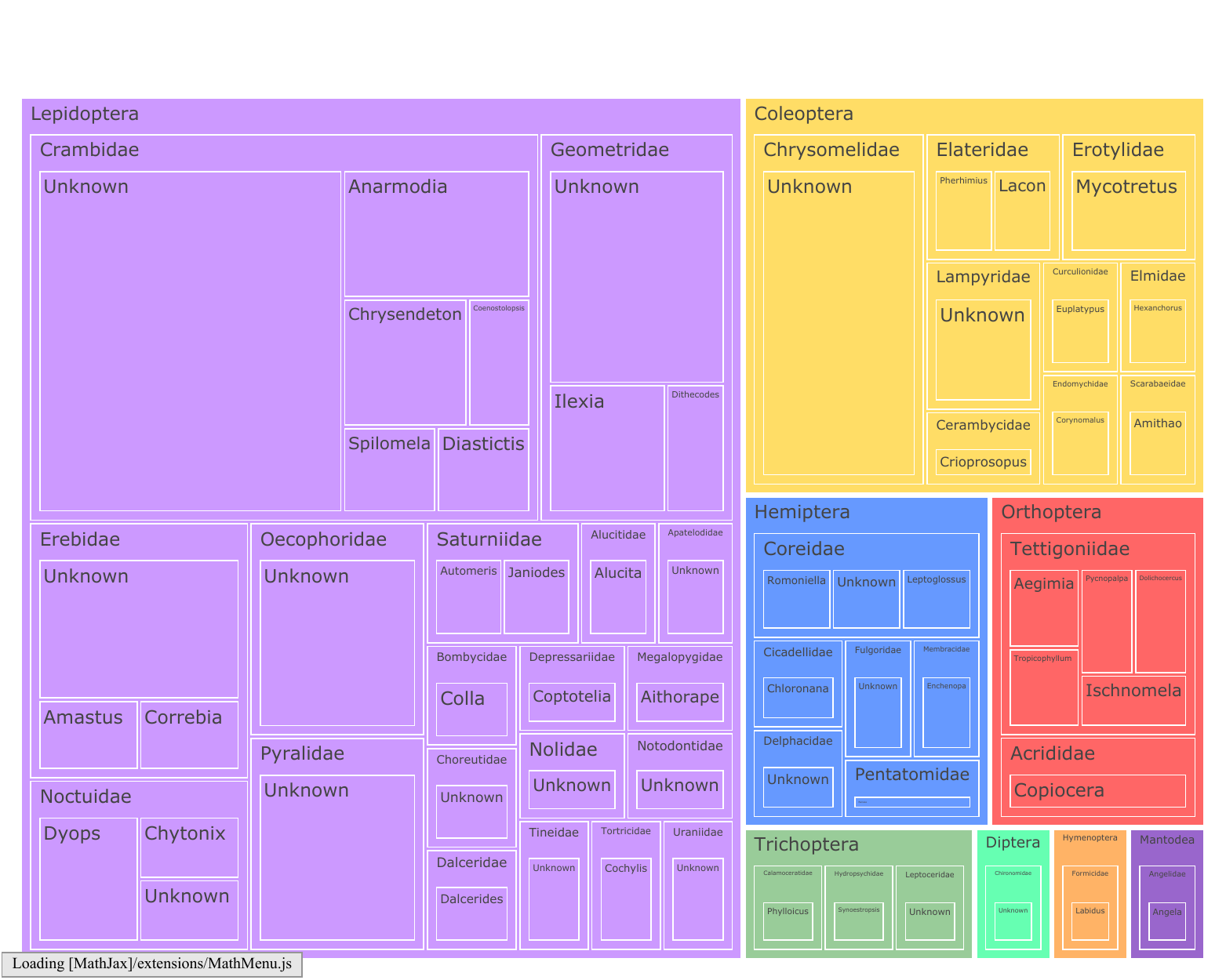} % or .pdf if you have it
    \caption{Treemap showing the hierarchical taxonomic distribution of \textit{Novel} insect species across Order, Family, and Genus levels. Box sizes are proportional to image counts.}
    \label{fig:new-treemap}
\end{figure}

\section{Introduction}

Biodiversity is vanishing faster than scientists can document it. Insects, the planet’s most species‑rich group, are declining at alarming rates, with many taxa disappearing before they are even described~\cite{smithsonian}. More than 80\% of insect species on Earth remain unnamed, despite playing pivotal roles in ecosystem stability, agriculture, and climate resilience~\cite{stork2024can}. Terry Erwin’s iconic canopy‑fogging experiment illustrates the magnitude of this hidden diversity: by fogging a single \textit{Luehea seemannii} tree in Panama, he recovered over 1,200 beetle species (163  specific to that tree)~\cite{Erwin1982,SmithsonianElNino}. But species discovery still relies on painstaking and expert‑led data acquisition and visual inspection, a workflow that is slow, subjective, and expensive.  Advances in high-throughput imaging technologies, such as those developed through the XPRIZE Rainforest competition~\cite{xprize_2024}, can revolutionize biodiversity surveys in complex ecosystems. Our paper includes contributions from members of the competition-winning team~\cite{limelight_team,wfu_magazine_2025}, who routinely leverage such techniques. However, despite high-throughput image acquisition, subsequent evaluation and scrutiny of these images remain reliant on expert judgment, and constrained by the alarmingly fast dwindling of expertise~\cite{IUCN2022Insects}

Recent progress in frontier vision–language models (VLMs) offers a promising alternative. Models trained on web‑scale multimodal corpora already excel at zero‑shot classification, cross‑domain transfer, and language‑guided reasoning, injecting objectivity and consistency into species identification while being able to process thousands of images per hour. Biodiversity‑focused datasets and models such as BioCLIP, BioTrove, and CLIBD highlight this potential~\cite{bioclip,biotrove,clibd}. Yet, reliable \emph{evaluation} of these models remains elusive: unlike math or code (where responses are verifiable), biodiversity data are noisy, long‑tailed, and inherently open‑world. Existing resources (like iNat2018, GBIF, iNat2021) have almost exclusively focus on \emph{closed‑set} recognition of known species~\cite{inat2018,gbif,inat2021}, but these reveal little about a model’s ability to detect \emph{unseen} taxa or justify predictions biologically, both central to ecological discovery~\cite{openset}.

Building upon this intersection of automated imaging and expert-led validation, we introduce~\textbf{TerraIncognita}, a unique biodiversity benchmark built with expert entomologists to test VLMs under realistic field conditions. The dataset blends research‑grade iNaturalist images, labelled down to species, with newly captured {images of rare and \emph{potentially novel} species that the latest models will have had no opportunity to train on}. These images were gathered through field campaigns in biodiversity hotspots across Central and South America.  For {most of these images}, only higher‑level labels are reliable, and many taxa are plausibly entirely new to science.  We query frontier models to output the full taxonomic hierarchy (Order $\rightarrow$ Family $\rightarrow$ Genus $\rightarrow$ Species) and return \texttt{Unknown} whenever finer‑level certainty is lacking. Evaluation simultaneously measures discovery accuracy, hierarchical accuracy, reliability when encountering novelty, and agreement between model explanations and expert diagnostic traits. We believe that TerraIncognita provides the first rigorous test‑bed for assessing how well frontier VLMs can move from classifying the known to discovering the unknown in the context of biodiversity. 

With this manuscript, we release the initial version of the dataset, present a comprehensive baseline study on state‑of‑the‑art models, and analyze whether their textual justifications are biologically plausible. To ensure the benchmark remains challenging and contamination-limited, we commit to quarterly updates. In each update cycle, we will add newly collected specimens--sourced from recent biodiversity campaigns--and retire outdated or overly familiar examples. On average, one-fourth of the dataset will be refreshed every cycle, enabling full turnover annually. Specimen selection for updates will prioritize the oldest entries and those that have become trivially solvable, thereby maintaining the scientific and diagnostic relevance of the benchmark over time.

\begin{itemize}[leftmargin=*,nosep]
    \item We introduce \textbf{TerraIncognita}, a benchmark dataset designed to test VLMs on insect biodiversity discovery across both known and potentially novel species. 
    \item We systematically evaluate state-of-the-art VLMs across both known and unknown insect taxa that reflect both hierarchical taxonomic accuracy and model reliability in uncertainty handling, as well as a structured explanation comparison against expert-sourced references. We highlight current limitations and strengths in several fine-grained open-set classification and OOD detection capabilities in frontier models.
    \item In addition to quantitatively benchmarking their performance, we qualitatively assess whether models generate biologically plausible explanations for their predictions. 
    \item We commit to \textbf{quarterly expansions}, including likely novel species, over the next year, enabling longitudinal benchmarking of model performance.
\end{itemize}

\section{The TerraIncognita Dataset}
The TerraIncognita test dataset is a dynamic benchmark designed to test VLMs in real-world biodiversity discovery scenarios. The dataset comprises approximately 200 insect species-level specimens split across two subsets of discovery categories. We will refer to these two subsets as \textit{Known} and \textit{Novel} for the remainder of this paper.
%—one in a standard dorsal or lateral pose and another in a non-standard view (e.g., side-angle, in motion, or partially occluded)—to reflect practical challenges encountered in field data
Figure~\ref{fig:fullpage} illustrates the complete data curation and collection pipeline, with a primary focus on the \textit{Novel} data acquisition. The first step shows our field expert photographing insects at a trap site, followed by a meticulous inspection and labeling process.

\textit{Novel} species images originate from light-trap images captured in the Ecuadorian Amazon and other unexplored biodiversity hotspots by expert field entomologist Dr. Harold F.\ Greeney, Director of Research at the Yanayacu Biological Station and Center for Creative Studies \cite{YanayacuStation}. These specimens were collected in remote locations in Central and South America and manually filtered and labelled by field experts. Unknown species have coarse-scale taxonomic labels (often limited to Order or Family, but no finer) and were intentionally selected from underrepresented or potentially novel taxa. Each specimen in \textit{Novel} is represented by at least two images. This yields a total of approximately 237 images with an average image resolution of approximately $2048 \times 1975$ pixels.

\textit{Known} species images are sourced (via visual inspection) from iNaturalist entries labeled as "research grade", filtered to match the taxonomic orders and families present in the \textit{Novel} set to mitigate label imbalance. These species have full taxonomic labels (Order $\rightarrow$ Family $\rightarrow$ Genus $\rightarrow$ Species), and image quality was ensured by selecting specimens with clearly visible anatomical features. Known species span over 8 insect orders and more than 20 families, with an average image resolution of approximately 1860$\times$ 1488 pixels comparable to the resolution of novel species images to ensure fair visual parity and avoid introducing bias due to image quality. Figure~\ref{fig:vlm-grid-orders} showcases images from both collections, spanning eight insect orders. The consistent quality and background across images help minimize bias during model evaluation.

The hierarchical distribution of novel insect species across Order, Family, and Genus levels is visualized in Figure \ref{fig:new-treemap}, where box sizes reflect the number of images per taxonomic group. Since our data is sourced from specific locations, the taxonomic distribution of novel insect species is skewed, with Lepidoptera dominating across all levels (also see Figure 1 in supplementary info).

\section{Data Collection and Curation Methodology}

%Our dataset comprises established species sourced from citizen-science images on iNaturalist~\cite{inaturalist} and potentially novel insect species collected directly from biodiversity-rich field locations. 
%This dual-source strategy uniquely tests the capabilities of general-purpose VLMs in recognizing both known and unfamiliar taxa.
\paragraph{Images of Known Species}
Known species images were manually selected from iNaturalist~\cite{inat2018}, adhering to three criteria:
%\begin{itemize}[leftmargin=*,noitemsep, topsep=0pt]
    %\item 
(i) Status: community-vetted (\textit{research grade}) with location tags; 
    %\item 
(ii) Taxonomic hierarchy: clearly labeled across Order, Family, Genus, and Species; and 
    %\item 
(iii) Visual quality: clear images with minimal background clutter.
%\end{itemize}
Species selection prioritized Orders (e.g., Lepidoptera, Hemiptera, Coleoptera, Hymenoptera) to closely match the novel species distribution.

\paragraph{Images of Novel Species}

{Images were collected by Dr.\ Harold Greeney's team, working primarily in northern Brazil, Ecuador, and Panama, across a range of habitats including tropical lowland and foothill forests, as well as subtropical montane cloud forests. These regions are well established as global biodiversity hotspots, particularly for Lepidoptera~\cite{Brehm_2005}. %The image collection process was informal and did not involve standardized stations or fixed camera setups. 
Most photographs were taken using a DSLR camera with a macro lens, occasionally supplemented with higher-end cameras, though a substantial number were captured using a smartphone. To attract nocturnal insects, a light source emitting wavelengths known to be particularly effective for Lepidoptera was used ~\cite{Brehm_2017, Brehm_2021}. The light was positioned in front of a white sheet or shower curtain to minimize background clutter.} Experts tagged unknown species typically to the Family level, with partial labels (primarily Order or Family) assigned to potentially novel species. Table~\ref{tab:dataset_summary} summarizes available taxonomic labels, and Table~\ref{tab:datacard} details the image metadata schema.\footnote{A key point of clarification is necessary regarding the term ``novel species''. In the biological sciences, declaring a species as ``novel'' typically implies that it has never been encountered or analyzed in the totality of scientific literature or discourse; essentially, it is ``new to science.'' Such claims are subject to rigorous scrutiny, and verifying this rigorously is outside the scope of our work. In the context of this paper, we do not intend to make such assertions. Instead, ``novel'' refers specifically to species that are unseen by the model --- that is, species that are very rare, and for which almost surely few (or no) labeled images exist in the model’s training data. It is entirely possible that other expert entomologists may be able to identify the taxonomy of many of these species in a zero-shot manner, and in that sense they are not necessarily ``new to science''. We have added clarifying language throughout the manuscript to emphasize that our evaluation focuses on the \emph{model's} ability to generalize to images that it has not encountered during training.} %, rather than making claims about scientific novelty.

\begin{table}[t]
\centering
\caption{Summary statistics for the TerraIncognita dataset split by known and unknown categories. Each entry reflects the number of images, labeled taxonomic levels, and species coverage. Known species have full taxonomic labels, while unknowns include partial hierarchy only.}
\label{tab:dataset_summary}
\begin{tabular}{lcccccc}
\toprule
\textbf{Category} & \textbf{\#Species} & \textbf{\#Images} & \textbf{\#Orders} & \textbf{\#Families} & \textbf{\#Genus} & \textbf{\#Species (Labeled)} \\
\midrule
Known     & 100 & 200 & 8 & 24 & 90 & 100 \\
Unknown   & 100 & 237 & 8 & 42  & 43 & 12 \\
\bottomrule
\end{tabular}
\end{table}

\begin{table}[t]
\caption{Representative entries from the TerraIncognita dataset card showing taxonomic labels and image views. Unknown entries may include only partial labels (e.g., up to Order or Family).}
\centering
\resizebox{0.8\textwidth}{!}{%
\begin{tabular}{@{}llccccl@{}}
\toprule
\textbf{Image ID} & \textbf{Status} & \textbf{Order} & \textbf{Family} & \textbf{Genus} & \textbf{Species}  \\
\midrule
IMG\_001 & Known   & Lepidoptera & Noctuidae     & Acronicta  & cyanescens \\
IMG\_002 & Known   & Hemiptera   & Reduviidae    & Zelus      & longipes \\
IMG\_003 & Unknown & Coleoptera  & Scarabaeidae  & NA         & NA\\
IMG\_004 & Unknown & Hymenoptera & Ichneumonidae & NA         & NA\\
\bottomrule
\end{tabular}}
\label{tab:datacard}
\end{table}

\subsection{Dataset Access and Quarterly Expansion Plan}
To ensure open access and reproducibility, the TerraIncognita benchmark dataset is publicly hosted on the Hugging Face Datasets Hub \footnote{\url{https://huggingface.co/datasets/BGLab/TerraIncognita}}. The dataset follows the Hugging Face datasets format and includes taxonomic metadata fields, along with the specimen's discovery status (Known or Unknown). Each image is paired with a structured metadata entry as illustrated in Table~\ref{tab:datacard}, which summarizes the format and available taxonomic depth per sample.

In addition to the current release, the benchmark will undergo quarterly updates, replacing approximately 25\% of the dataset each cycle with newly collected specimens to ensure novelty, minimize contamination, and maintain long-term challenge integrity. We commit to maintaining this schedule for at least one year. This will offer an opportunity to perform longitudinal tracking in the evolution of frontier models in their ability to perform open-world visual recognition.

\begin{figure}[hbtp]
    \centering
    \includegraphics[width=\textwidth]{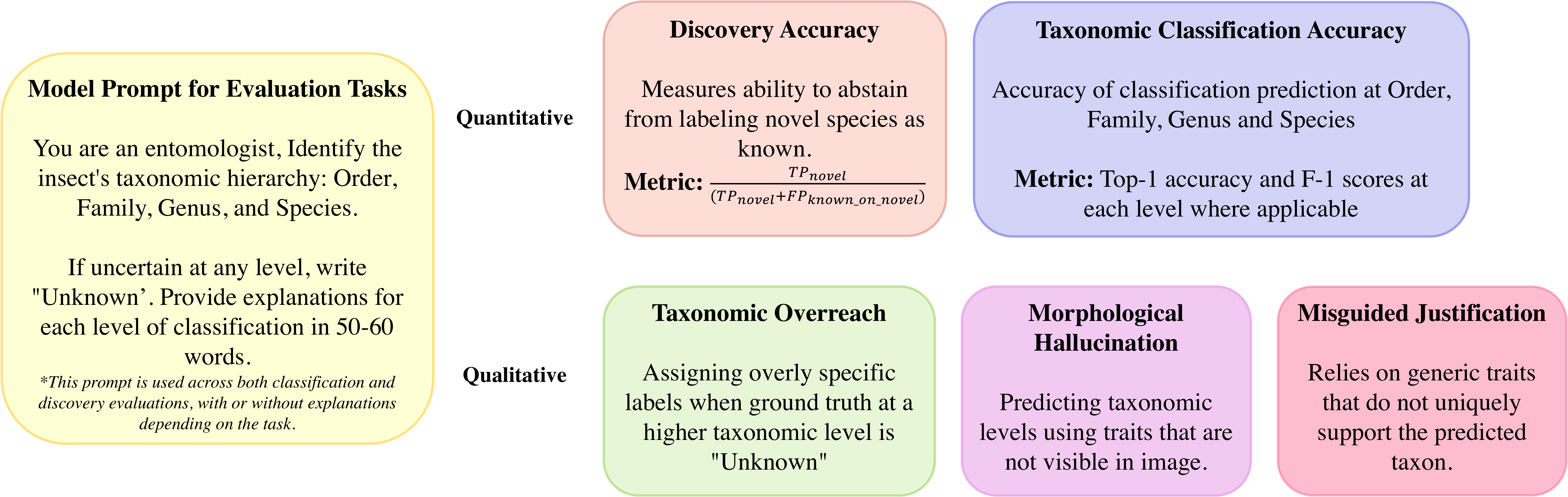}
    \caption{
    \textbf{Evaluation Dimensions for Open-World Insect Classification.} 
    We evaluate models along both quantitative and qualitative dimensions using a shared entomology-based prompt. This prompt asks the model to identify insect taxonomy at four levels --- Order, Family, Genus, Species --- abstaining when uncertain. Discovery accuracy measures the model's ability to correctly abstain on novel species, while classification accuracy quantifies correctness at each taxonomic level. Taxonomic overreach and morphological hallucination among others capture failure modes in open-world reasoning. Explanation alignment assesses whether model-generated justifications correspond with expert-reviewed traits.
    }
    \label{fig:evaluation-taxonomy}
\end{figure}
\section{Models and Benchmarks}

We evaluate TerraIncognita on 12 state-of-the-art VLMs, encompassing a range of architectures, training corpora, and modalities. These include: GPT-4.1, GPT-4o, o3, Claude variants (Claude-3-Opus, Claude-3.5-Haiku, Claude-3.7-Sonnet), Gemini 2.5 Flash, Grok-2-Vision, Qwen (Qwen 2VL and Qwen2.5 VL), and LLaMA-4 variants (Scout and Maverick). All models were accessed via their APIs, with image inputs provided in raw format.

Each model was evaluated under a unified prompting framework. Specifically, we used a fixed system prompt directing the model to classify an insect specimen across the taxonomic hierarchy (Order → Family → Genus → Species), returning “Unknown” when uncertain which is illustrated in Figure 5. No few-shot exemplars or retrieval augmentations were used, ensuring pure zero-shot evaluation. This controlled setup allows us to isolate differences in model representations and reasoning capacity, rather than prompt engineering or fine-tuning artifacts. All prompts are detailed in the Supplementary material.

\subsection{Evaluation Metrics}

We evaluate VLMs across two discovery settings: \textit{Known} species, where full taxonomic labels are available, and \textit{Novel} species, where partial labels are present where abstention is the desired behavior. We evaluate model performance using quantitative metrics and qualitative assessments. The evaluation framework encompasses both structured accuracy scores and expert-informed analyses across key taxonomic prediction tasks. The quantitative metrics used are as follows.
\begin{itemize}[topsep=2pt, itemsep=2pt, parsep=0pt, leftmargin=1.5em]
    \item \textbf{Discovery accuracy.} This binary metric captures whether the model correctly distinguishes between \textit{Known} and \textit{Novel} species. For \textit{Known} images, a correct prediction implies confident output at the species level. For \textit{Novel} images, the correct behavior is abstention (i.e., predicting ``Unknown'') at the Species or higher levels, depending on label availability.
    \item \textbf{Taxonomic accuracy.} We compute accuracy at each taxonomic level (Order, Family, Genus, and Species) for \textit{Known} examples. This evaluates the model's ability to match ground truth labels at increasing levels of specificity. For \textit{Novel} species, we evaluate model predictions only at the Order and Family levels, which are the taxonomic ranks for which expert-provided ground truth labels are available. 
\end{itemize}
% \newline \newline
% \textbf{Hierarchy-Aware Accuracy.} As an optional metric, we include taxonomic distance: a penalty-based score based on how far the model's prediction deviates from the correct node in the taxonomic tree. Predicting the wrong Family within the correct Order is penalized less than predicting a completely wrong Order. This metric is useful in cases where exact matches are too strict.
% (Explanation:(makes more sense for known species as depth is 4, compared to unknown where depth is 1  or 2) Each entity is a leaf node in a taxonomy tree, and predictions are evaluated by summing the weights of shared ancestor nodes between the predicted and true paths—from root to leaf.
% Weights decrease with depth (e.g., Order = 0.5, Family = 0.25, Genus = 0.125), so matches at broader levels contribute more to the score than finer-level matches (can do it vice verse).
% This makes the metric especially useful for cases with incomplete labels (e.g., labeled only up to Order or Genus), as it allows partial credit and avoids discarding examples where .)
% \newline \newline
To better understand model behavior, we conducted a qualitative analysis using both the taxonomic labels and the corresponding textual explanations produced by VLMs at each taxonomic level (order, family, genus, species). These predictions and explanations are reviewed by an expert entomologist to identify common reasoning failures.

Rather than scoring explanation quality directly, we used the explanations as a lens to categorize recurring failure modes, capturing how and why models arrive at incorrect predictions. This analysis revealed systematic reasoning errors, ranging from superficial visual matching to biologically implausible trait references. Representative examples for each failure mode are shown in Table ~\ref{tab:failure_modes_categorized}.

\begin{itemize}[topsep=2pt, itemsep=2pt, parsep=0pt, leftmargin=1.5em]
\item \textbf{Morphological hallucination:} The model references traits that are not visible in the image—e.g., internal structures or obscured morphology.
\item \textbf{Speculative inference:} Justifications are based on behavioral, ecological, or lifecycle traits that cannot be inferred from the image alone.
\item \textbf{Misguided justification:} The model cites real but overly generic traits that are not diagnostic of the predicted taxon.
\item \textbf{Style over substance:} The model relies on superficial visual impressions such as pose, outline, or “vibe” rather than taxonomically meaningful traits.
\item \textbf{Taxonomic overreach:} The model commits to an overly fine-grained prediction (e.g., species) without sufficient visual evidence, often against expert judgment.
\end{itemize}

This explanation-guided review provides insight into whether model predictions fail due to missing key morphological signals, misinterpreting ambiguous features, or hallucinating biologically unsupported justifications. Table~\ref{tab:failure_modes_categorized} presents illustrative examples across different taxonomic levels. Additional cases are provided in the Supplementary Information section.

\begin{table}[t]
\caption{Discovery Accuracy comparison between \textit{Known} and \textit{Novel} species across models.}
\label{tab:discovery_accuracy}
\centering
\begin{tabular}{lccc}
\toprule
\textbf{Model} & \textbf{Known Disc. Acc.} & \textbf{Novel Disc. Acc.} & \textbf{F1.} \\
\midrule
GPT-4o            & 16.50  & 77.64 & 66.36\\
GPT-4.1           & 61.00  & 55.27 & 73.33\\
o3           & 31.50  & 80.17 & 73.33\\
Claude-3-Opus     & 0.00   & 87.76 & 64.50\\
Claude-3.5-Haiku  & 31.50  & 75.11 & 71.09\\
Claude-3.7-Sonnet & 47.00  & 67.93 & 73.70\\
Gemini-2.5-Flash  & 40.00  & 58.65 & 66.77\\
LLaMA-4-Scout     & 24.50  & 81.43 & 71.28\\
LLaMA-4-Maverick  & 5.00   & 87.76 & 66.56\\
Grok-2-Vision     & 1.00   & 87.76 & 64.91\\
Qwen2-VL          & 16.00  & 75.53 & 65.12\\
Qwen2.5-VL        & 19.00  & 75.11 & 66.16\\
\bottomrule
\end{tabular}
\end{table}

\begin{table}[t]
\caption{Performance of VLMs on the biodiversity benchmark. We report taxonomic metrics across all taxonomic levels for \textit{Known} species and at the Order and Family levels for \textit{Novel} species.}
\label{tab:model_results}
\centering
\resizebox{0.8\textwidth}{!}{%
\begin{tabular}{@{}llccccc@{}}
\toprule
\textbf{Species Type} & \textbf{Model}  & \textbf{Ord. F1} & \textbf{Fam. F1} & \textbf{Gen. F1} & \textbf{Spec. F1} \\
\midrule
\multirow{14}{*}{\textbf{\textit{Known}}} 
& GPT-4o                & 88.85 & 40.76 & 5.59 & 0.39\\
& GPT-4.1               & 100 & 44.26 & 7.97 & 0.0\\
& o3               & 100 & 44.19 & 13.69 & 1.36\\
& Claude-3-Opus         & 52.83 & 26.11 & 1.20 & 0.0\\
& Claude-3.5-Haiku      & 80.09 & 35.89 & 5.96 & 0.45\\
& Claude-3.7-Sonnet     & 88.85 & 36.64 & 7.89 & 1.01\\
%& Gemini-1.5-Flash     & 2.00 & 100.0 & 42.97 & 0.00 & 0.00\\
& Gemini-2.5-Flash      & 88.78 & 38.57 & 7.77 & 3.00\\
%& Gemini-2.5-Pro       & 60.00 & 88.85 & 42.09 & 11.48 & 5.49\\
& LLaMA-4-Scout         & 66.38 & 23.36 & 1.28 & 0.00\\
& LLaMA-4-Maverick      & 43.28 & 28.56 & 1.13 & 0.00\\
& Grok-2-Vision         & 84.07 & 27.37 & 1.74 & 0.99\\
& Qwen2-VL              & 91.59 & 30.86 & 2.42 & 0.52\\
& Qwen2.5-VL            & 74.29 & 42.69 & 11.79 & 1.47\\
\midrule
\multirow{14}{*}{\textbf{\textit{Novel}}} 
& GPT-4o                & 38.5 & 15.40 & -- & -- \\
& GPT-4.1               & 49.5 & 16.66 & -- & -- \\
& o3               & 52.98 & 15.81 & -- & -- \\
& Claude-3-Opus         & 24.14 & 3.81 & -- & -- \\
& Claude-3.5-Haiku      & 39.21 & 7.29 & -- & -- \\
& Claude-3.7-Sonnet     & 47.24 & 13.90 & -- & -- \\
%& Gemini-1.5-Flash     &        &      &       & -- & -- \\
& Gemini-2.5-Flash       & 41.55 & 15.50  & -- & -- \\
%& Gemini-2.5-Pro       &        &      &      & -- & -- \\
& LLaMA-4-Scout         & 45.37 & 9.36 & -- & -- \\
& LLaMA-4-Maverick      & 28.91 & 10.89 & -- & -- \\
& Grok-2-Vision         & 45.36 & 11.49 & -- & -- \\
& Qwen2-VL              & 41.33 & 10.75 & -- & -- \\
& Qwen2.5-VL            & 35.49 & 11.78 & -- & -- \\
\bottomrule
\end{tabular}%
}
\end{table}
\section{Experimental Results}

%\subsection{Quantitative Performance}

\textbf{\textit{Known} Species}: Order-level performance is near-perfect for many models (e.g., GPT-4.1, o3, Qwen 2VL), with F1 scores exceeding 90\% (refer Table \ref{tab:model_results}). However, Family- and Genus-level F1 scores decline sharply, with only a handful of models—such as o3 and Qwen 2.5 VL—achieving Genus F1 >10\%. Species-level predictions remain low across the board, reflecting the inherent difficulty of fine-grained entomological classification. For example, Qwen 2.5 VL achieves 1.47\% Species F1, while o3 peaks at 1.36\%. Importantly, this performance is achieved under a strict zero-shot setting with no exposure to training taxonomies or in-domain examples. While the known species dataset is unbalanced and heavily skewed toward the Lepidoptera order (to mirror \textit{Novel} species), the models had no prior knowledge of this distribution—yet still demonstrated strong generalization at higher taxonomic levels. This highlights both the promise and the limitations of current VLMs under open-world biodiversity constraints, where coarse-level reasoning remains feasible but fine-grained distinctions are far more elusive.
We also evaluate discovery accuracy in Table ~\ref{tab:discovery_accuracy}, defined as the ability to correctly not abstain on known examples. Surprisingly, models struggle even here: only GPT-4.1 exceeds 60\% discovery accuracy, with most other models doing poorly despite the presence of large scale publicly available datasets like iNaturalist (refer Table \ref{tab:discovery_accuracy}). This suggests that VLMs often default to uncertainty ("Unknown") even when the taxonomy is fully known, highlighting a conservative bias and an inability to calibrate their predictions based on visual confidence alone. In all, frontier models continue to exhibit key gaps in zero-shot biodiversity reasoning and fine-grained image classification.

\textbf{\textit{Novel} Species}:
Evaluation on novel insect specimens presents a distinct set of challenges. Unlike the known species set, where ground truth spans all four taxonomic levels, labels for novel species are partial, typically limited to Order and Family. Consequently, models are assessed based on two key dimensions: their ability to abstain from overconfident predictions (captured by discovery accuracy) and their coarse-level taxonomic classification ability (via Order and Family F1 scores). Performance across models reflects the difficulty of open-world classification. Discovery accuracy is highest for Claude-3-Opus, Grok-2-Vision, and LLaMA-4-Maverick, all exceeding 87\% (ref. Table ~\ref{tab:discovery_accuracy}. These results suggest that some VLMs have learned to adopt conservative prediction strategies when faced with novel taxa, abstaining appropriately instead of hallucinating fine-grained classifications. In contrast, other models (e.g., GPT-4.1, Gemini-2.5 Flash) show lower discovery accuracy, indicative of premature overcommitment at the Genus or Species level despite lacking strong visual or contextual evidence. F1 scores at the order level remain moderate across the board, generally ranging from 24\% to 52\% as seen in Table ~\ref{tab:model_results}. While this suggests some degree of generalization to unseen taxa, it also reflects the difficulty of assigning even coarse taxonomic labels to novel inputs. Family-level F1 scores are notably lower, often dropping below 20\%, underscoring the challenge of fine-grained reasoning in an open-set setting without species-level supervision.

\textbf{Comparative trends between models}: We make the following observations:
\begin{enumerate}[leftmargin=*,noitemsep, topsep=0pt]
    \item OpenAI's GPT series shows the most consistent behavior across known/unknown categories.
    \item Google's Gemini 2.5 models achieves higher than average Order/Family accuracy, but tend to over-commit on unknowns. 
    \item Claude and Qwen models abstain more aggressively but sacrifice finer-level specificity.
    \item Species-level predictions, though rare, are most successfully attempted by o3, and Qwen2.5VL Vision.
\end{enumerate}

%\subsection{Qualitative Analysis}

To complement our quantitative evaluations, we performed a qualitative analysis (see Supplementary Info) examining model-generated explanations at each taxonomic level. Expert entomologists reviewed these explanations to identify systematic reasoning errors such as morphological hallucinations, speculative inferences, and taxonomic overreach. 

%This analysis provides critical insights into the interpretability of model decisions, highlighting the extent to which predictions align with biologically plausible reasoning and revealing areas requiring further methodological improvement.

\section{Discussion and Limitations}

We conclude by providing a brief discussion on limitations, broader impact, and related work

\textbf{Related work}: Recent advances in biodiversity datasets ~\cite{bioclip,biotrove,bioscan,bioscan5m} have introduced multimodal resources combining images, DNA, and taxonomic metadata, enabling improved clustering and zero-shot classification. However, most operate under closed-set assumptions, limiting their utility in realistic open-world scenarios where novel taxa abound. In computer vision and vision-language models (VLMs), various strategies exist for open-set recognition (OSR) and out-of-distribution (OOD) detection exist ~\cite{msp,mojdeh,openset_deep,chenhui,danhendrycks,feedtwo,neco}. Yet, VLMs remain biased toward known categories, often misclassifying novel inputs. Biodiversity-focused open world dataset efforts (e.g., InsectSet459, Open-Insects) offer partial solutions but often rely on simulated novelty ~\cite{openset_insects,insect459}. As detailed in Section~\ref{sec:SI_related_work}, TerraIncognita addresses these limitations by providing real-world, expert-vetted insect images that challenge VLMs to perform fine-grained, biologically grounded classification under true open-world conditions.

\textbf{Challenges in \textit{Novel} species assessment.} While VLMs show promising coarse-grained taxonomic reasoning in known insect classification (e.g., order-level accuracy >90\% for several models), their performance deteriorates steeply for \textit{Novel} species, particularly beyond the Family level. This aligns with known difficulties in open-world biodiversity discovery, where many specimens belong to cryptic or underrepresented taxa. We find a consistent pattern of overcommitment in unknown cases, where models prematurely assign Genus or Species labels despite lacking visual confidence or training coverage. This suggests that current VLMs lack good mechanisms for abstention, which is a critical trait for AI-based decision-making that extends beyond biodiversity applications.

\textbf{Interpretability as a bridge to expert trust.} Our qualitative analysis shows that there is a discrepancy between model-generated explanations and expert reasoning (see Table \ref{tab:failure_modes_categorized}) While some responses highlighted plausible morphological cues, others hallucinated features or descriptors. This indicates a need for biologically-grounded explanation benchmarks in addition to label-level metrics.

\textbf{Dataset scale and taxonomic breadth.} 
TerraIncognita comprises approximately 450 images covering 200 species across 8 Orders. While modest in size, its curated diversity offers a potent testbed for evaluating vision–language models (VLMs) in real-world biodiversity contexts. The inclusion of rare and potentially novel species challenges models to generalize beyond familiar taxa. Although the limited scale may affect statistical power and introduce biases toward over-represented groups like \textit{Lepidoptera}, the dataset's design facilitates focused assessments of model capabilities. 
% Although TerraIncognita is diverse, it remains modest in scale: ~437 images spanning 200 species across 8 Orders. This limits the statistical power of some comparisons and may bias models toward over-represented taxa (e.g., Lepidoptera). Moreover, \textit{Novel} species were labeled only up to Order or Family in most cases, restricting the evaluation depth. More diverse specimens and deeper annotations in the future iterations (perhaps using other modalities, such as DNA barcoding) would strengthen ground truth.

\textbf{Longitudinal benchmarking.} A major contribution of our benchmark is its dynamic nature. By committing to quarterly additions of 400+ new samples (including potentially undescribed species), we provide a framework for longitudinal evaluation. As VLMs improve or are fine-tuned on biodiversity datasets, TerraIncognita will track their evolution across both classification accuracy and interpretability metrics.

\textbf{Broader implications.} The real-world utility of AI in biodiversity depends not just on accuracy, but on conservatism, interpretability, and being  ``right for the right reasons''. Our findings suggest that while current frontier models offer a good baseline, they fall short of replacing expert taxonomists in fine-grained classification. Instead, we envision frontier models being used as triaging tools within a larger system, flagging likely novel specimens, prioritizing those requiring expert review, and generating biologically plausible explanations to aid experts.

\bibliographystyle{unsrt} 
\bibliography{references}

\section{Supplementary Information}

\subsection{Related Work}
\label{sec:SI_related_work}
In real-world biodiversity applications, models frequently encounter specimens that belong to species or taxa not present in the training data. Insects represent the majority of Earth’s biodiversity, yet a vast number of species remain undocumented or underrepresented in existing datasets~\cite{stork2024can,tax_bias}. Recent large-scale resources like BIOSCAN-1M/ BIOSCAN-5M and BioTrove~\cite{bioscan,bioscan5m,biotrove} have made important strides by introducing multimodal datasets that combine images, DNA barcodes, or taxonomic metadata for millions of insect specimens—enabling improved clustering and zero-shot classification performance. Similarly, contrastive learning approaches have been leveraged to align biological images with textual taxonomic labels or DNA barcodes as shown in BIOCLIP, CLIBD, BioTrove-CLIP etc.,~\cite{bioclip,clibd, biotrove} which trained a CLIP-style model on biological datasets for zero-shot classification. While these advances have accelerated biodiversity research, they largely operate under a closed-set assumption, where models expect all test instances to belong to known categories. However, this assumption breaks down in the wild, where many specimens encountered by monitoring systems may be undescribed or fall outside the training taxonomy. Existing datasets provide limited support for evaluating how well models handle such novel or out-of-distribution (OOD) instances or perform Open-Set Recognition (OSR)—highlighting the need for benchmarks that explicitly test these capabilities.

\textit{OSR and OOD in Computer Vision:} A diverse set of approaches have been developed to address the challenge of detecting unseen or novel classes—crucial for biodiversity monitoring in open-world conditions. Post-hoc methods, such as Maximum Softmax Probability (MSP)\cite{msp} among others ~\cite{neco,openset_deep,enhance_ood}, estimate confidence from trained models and remain widely used due to their simplicity. Energy-based models~\cite{mojdeh, weitang} estimate how likely an input is to come from the known training distribution by assigning it a scalar energy score—lower scores indicate in-distribution inputs, while higher scores suggest novelty. Methods that include training, include techniques like outlier exposure~\cite{danhendrycks} or outlier distribution adaptation~\cite{wenjun}, aim to proactively separate known and unknown classes during model optimization. Additionally, several methods incorporate external datasets during training to expose models to a broader diversity of classes or distributions~\cite{chenhui, feedtwo}, improving generalization under semantic shift. While these methods have been extensively evaluated on benchmarks with synthetic novelty (e.g., CIFAR-10 ~\cite{cifar10}),  they often fail to capture the real-world complexity inherent in biodiversity datasets, where novel classes may differ subtly from known ones and lack clear decision boundaries.

\textit{OSR and OOD in VLMs:} VLMs, despite being trained on internet-scale data, are not inherently open-set and exhibit closed-set biases due to their limited query sets. These models frequently misclassify OOD inputs, with increasing query sizes failing to improve precision-recall tradeoffs ~\cite{dimitymiller}. Recent benchmarks highlight their poor reliability under OOD conditions, particularly in visual question answering (VQA), where answer confidence is often a weak indicator ~\cite{xiangxishi}

\textit{OSR in Biodiversity Benchmarks:} Recent efforts in insect biodiversity monitoring span multiple modalities. InsectSet459 introduces the first large-scale acoustic dataset for insect sound classification, covering 459 \textit{Orthoptera} and \textit{Cicadidae} species with over 26,000 recordings—enabling deep learning methods for passive audio monitoring ~\cite{insect459}. Separately, a Bayesian open-set model combining images and DNA barcodes demonstrated strong performance in classifying known species and detecting unseen ones, highlighting the value of multimodal data for novel species discovery ~\cite{openset_insects1}.
In the visual domain, Open-Insects offers a fine-grained benchmark for open-set recognition, with geographically motivated OOD splits and a dedicated test set of likely undescribed species ~\cite{openset_insects}. Unlike smaller datasets like SSB-CUB or iNat21-OSR, it emphasizes ecological realism and semantic proximity in unseen class detection. However, these datasets rely heavily on simulated novelty, typically by holding out known species or constructing taxonomic hop-based splits within well-studied groups like birds or moths ~\cite{ssb_cub,inat_osr}. In contrast, TerraIncognita introduces real-world, field-collected images of likely undescribed insect species, spanning diverse orders and ecological contexts. Moreover, while existing benchmarks focus on post-hoc or regularized vision models, TerraIncognita uniquely targets VLMs, testing their capabilities in hierarchical classification, novelty detection, and taxonomic reasoning—with outputs aligned to expert ecological explanations.

\begin{figure}[h!]
    \centering
    \begin{subfigure}[t]{\textwidth}
        \centering
        \includegraphics[width=0.7\textwidth]{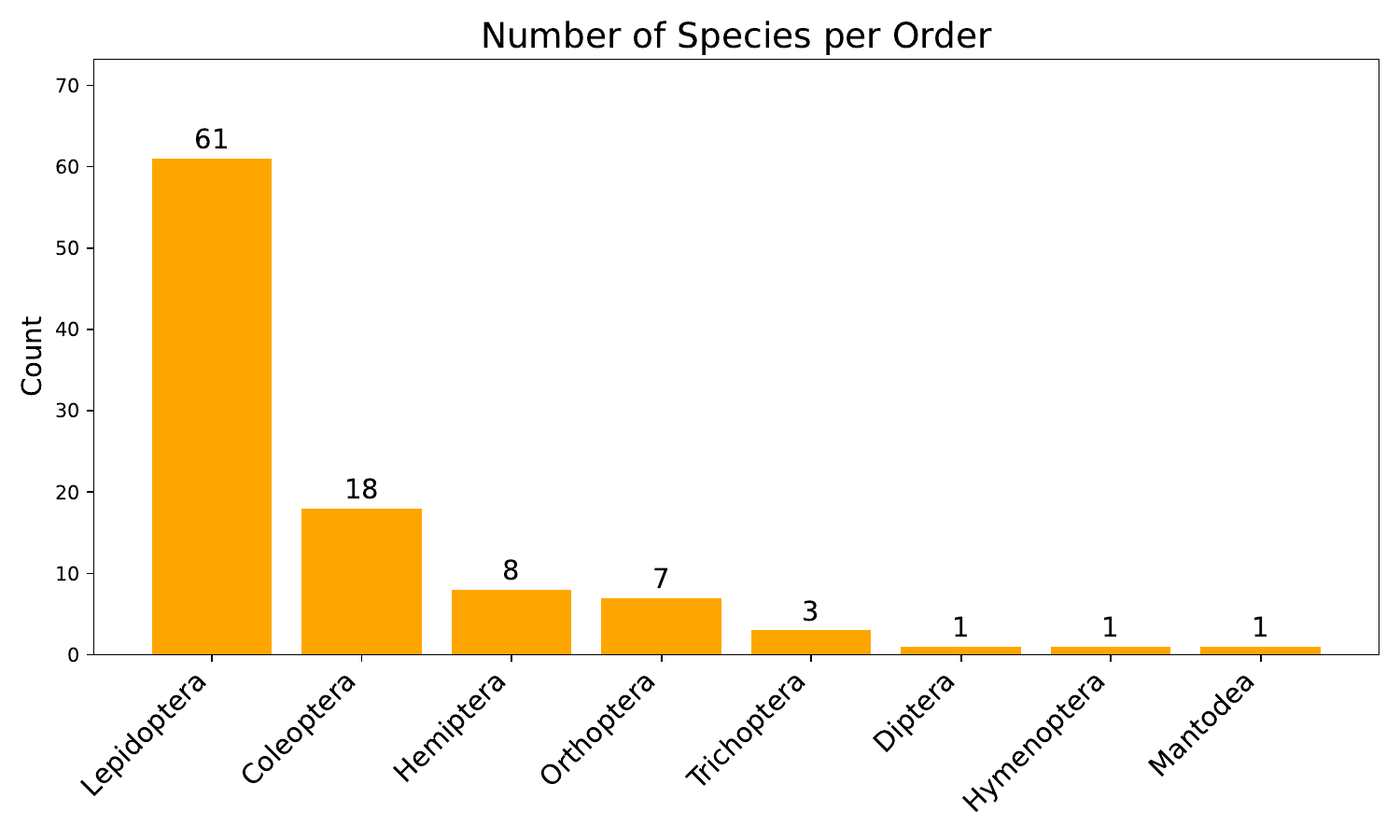}
        \caption{Number of species per insect order in new species set.}
        \label{fig:new-order-dist}
    \end{subfigure}

    \vspace{1em} % Adds spacing between the plots

    \begin{subfigure}[t]{\textwidth}
        \centering
        \includegraphics[width=\textwidth]{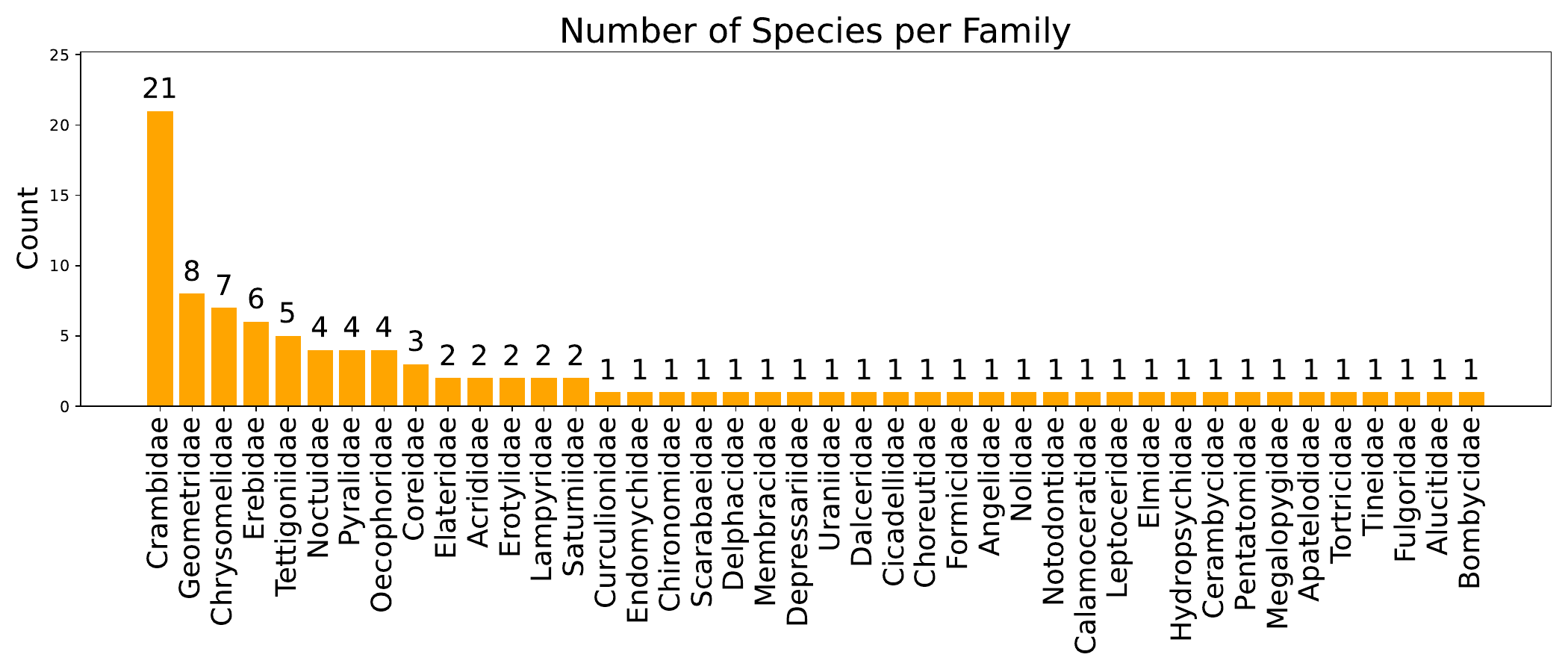}
        \caption{Number of species per insect family in new species set.}
        \label{fig:new-family-dist}
    \end{subfigure}
    \caption{Distribution of species by order and family for \textit{Novel} species. The dataset is skewed toward order of Lepidoptera and family Crambidae.}
    \label{fig:new-taxa-distribution}
\end{figure}

\subsubsection*{Species Distribution by Order}
As shown in Figure~\ref{fig:new-order-dist}, the dataset is heavily dominated by the order \textit{Lepidoptera}, which accounts for the largest number of species (n=61). Other frequently occurring orders include \textit{Coleoptera} (n=18), \textit{Hemiptera} (n=8), and \textit{Orthoptera} (n=7). A small number of species belong to less common orders in the dataset such as \textit{Mantodea, Diptera,} and \textit{Hymenoptera}. This skewed distribution reflects the sampling bias and relative richness of \textit{Lepidoptera} in the collected datasets.

\subsubsection*{Species Distribution by Family}
 Figure~\ref{fig:new-family-dist} shows the species counts at the family level. \textit{Crambidae} is the most represented family overall, with 21 species, followed by \textit{Geometridae} (8), \textit{Chrysomelidae} (7) and \textit{Erebidae} (6). Other top families include \textit{Tettigoniidae, Noctuidae, Pyralidae,} and \textit{Notodontidae}, reflecting a diversity of moth and beetle lineages.  However, the long tail includes many families represented by only one or two species, suggesting both a wide taxonomic spread and the potential for discovering novel or underrepresented families in insect biodiversity monitoring.

\subsection{Qualitative Insights}

\begin{table*}[ht]
\centering
\small
\renewcommand{\arraystretch}{1.4}
\resizebox{\textwidth}{!}{%
\begin{tabular}{|m{2.2cm}|m{3.0cm}|m{6.3cm}|m{6.3cm}|}
\hline
\textbf{Image} & \textbf{Taxonomic Level} & \textbf{Explanation} & \textbf{Failure Mode} \\
\hline

\multirow{4}{*}{\includegraphics[width=2.2cm]{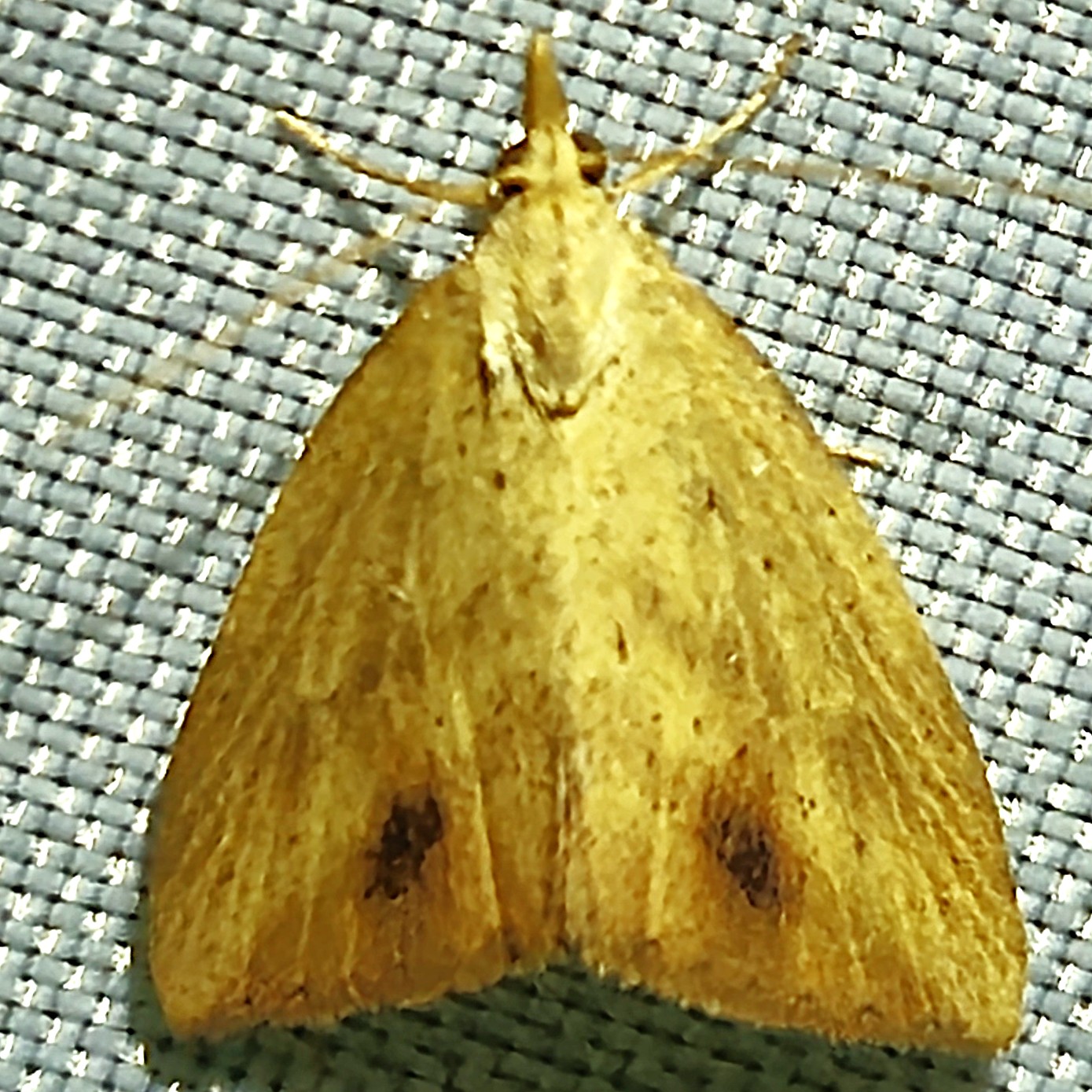}} 
& Order:\textit{Lepidoptera} \textcolor{green}{\ding{51}} & Scaled wings are visible, but coiled proboscis and sheen are not. 
& Style over Substance, Morphological Hallucination \\
& Family: \textit{Erebidae} \textcolor{green}{\ding{51}} & Traits (e.g., triangular stance, palps, antennae) are reused across families; not exclusive to Erebidae.
& Misguided Justification \\
& Genus: \textit{Rivula} \textcolor{red}{\ding{55}} & Superficial morphological match, but expert flags geographic mismatch and subtle spot pattern discrepancies.
& Misguided Justification \\
& Species: \textit{propinqualis}\textcolor{red}{\ding{55}} & Predicted species \textit{R. propinqualis} occurs $>$1000km north; expert suggests possible undescribed species.
& Taxonomic Overreach \\
\hline

\multirow{4}{*}{\includegraphics[width=2.2cm]{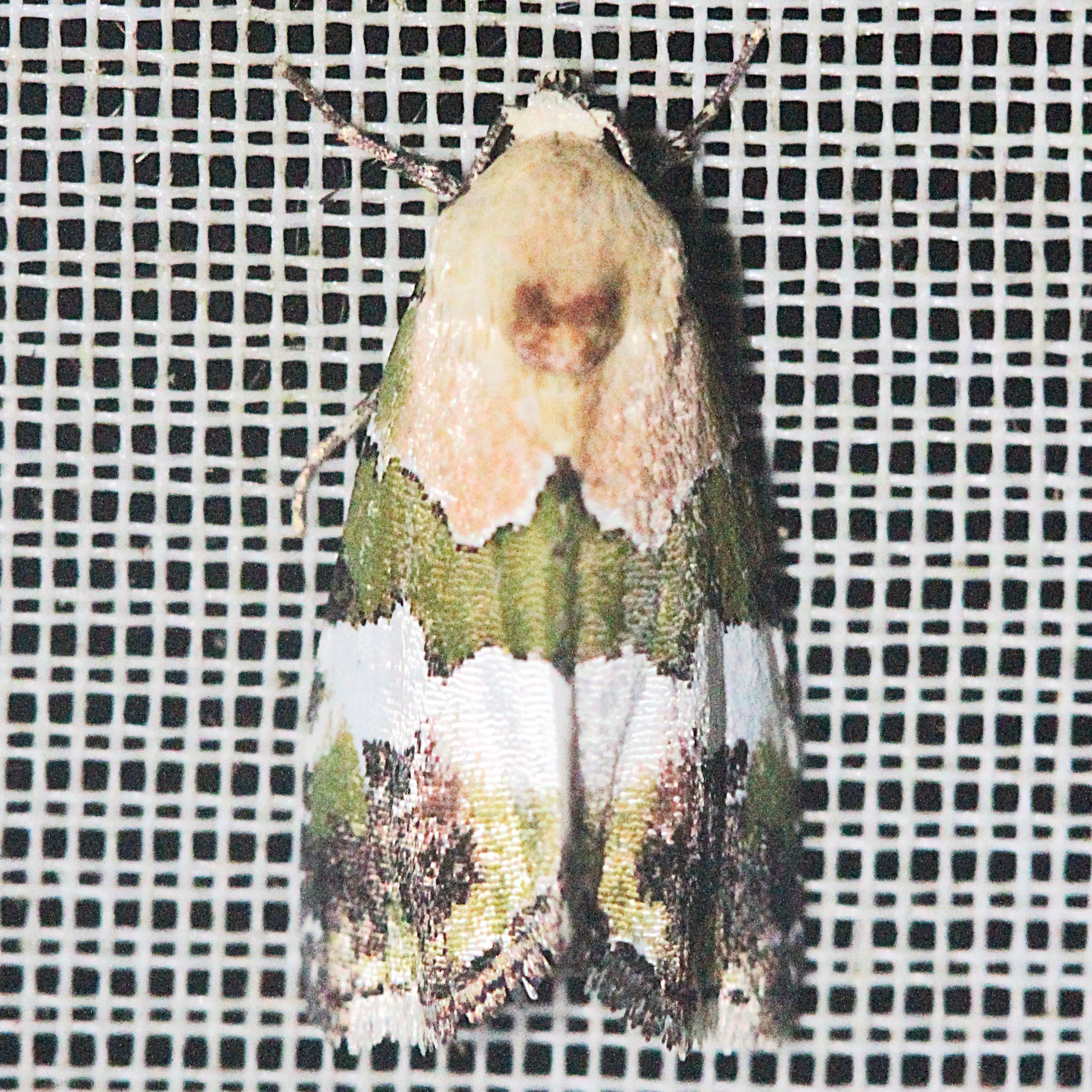}} 
& Order:\textit{Lepidoptera} \textcolor{green}{\ding{51}} & Cites standard \textit{Lepidoptera} traits (scaled wings, roof-like posture), but mentions coiled tongue and metamorphosis which aren't visible.
& Morphological Hallucination \\
& Family:\textit{Noctuidae} \textcolor{green}{\ding{51}} & Mentions tympanal organs (invisible in image) and references Acontiinae early, anchoring reasoning.
& Morphological Hallucination \\
& Genus:\textit{Ponometia} \textcolor{red}{\ding{55}} & Describes features like color zones and mimicry (e.g., bird-dropping resemblance), but expert says these can't be known from image.
& Speculative Inference \\
& Species: \textit{candefacta} \textcolor{red}{\ding{55}} & Assigned \textit{Ponometia candefacta} but expert says no match; proposes \textit{Elaphria} or undescribed species.
& Taxonomic Overreach \\
\hline

\multirow{4}{*}{\includegraphics[width=2.2cm]{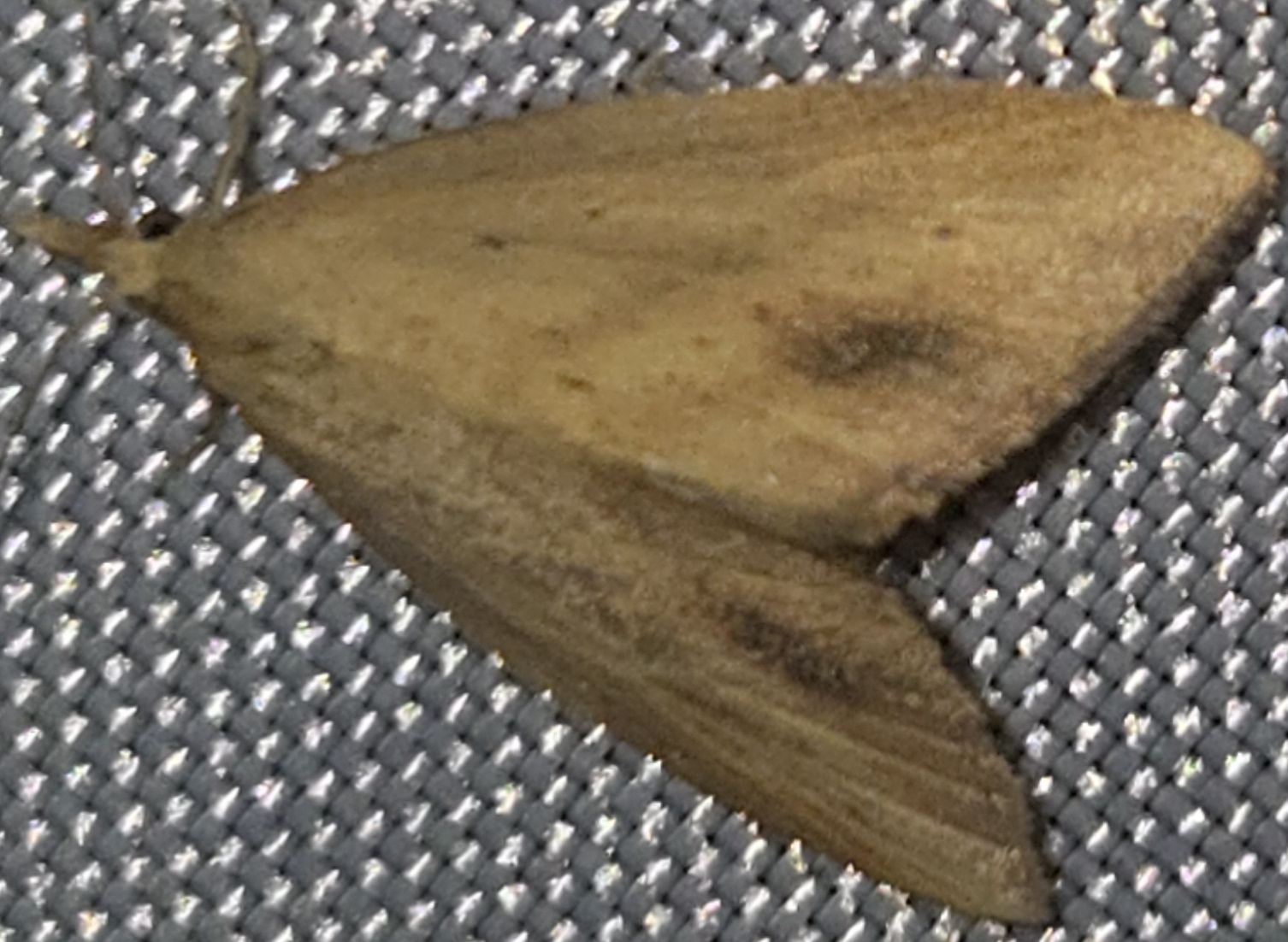}} 
& Order: \textit{Lepidoptera} \textcolor{green}{\ding{51}} & Mentions scaled wings and coiled proboscis; the latter not visible. Relies heavily on posture.
& Morphological Hallucination, Misguided Justification \\
& Family:\textit{Pyralidae} \textcolor{red}{\ding{55}} & Reuses generic traits (snout-like palps, narrow forewings) seen in many families, not diagnostic for Pyralidae.
& Style over Substance, Misguided Justification \\
& Genus: \textit{Achroia}  \textcolor{red}{\ding{55}} & Expert cannot trace any similarity to \textit{Achroia}; shape, posture, pattern all mismatch.
& Severe Mismatch \\
& Species: \textit{grisella} \textcolor{red}{\ding{55}} & Assigned \textit{Achroia grisella}, a common species, to what expert says is an unrelated specimen.
& Taxonomic Overreach \\
\hline

\multirow{4}{*}{\includegraphics[width=2.2cm]{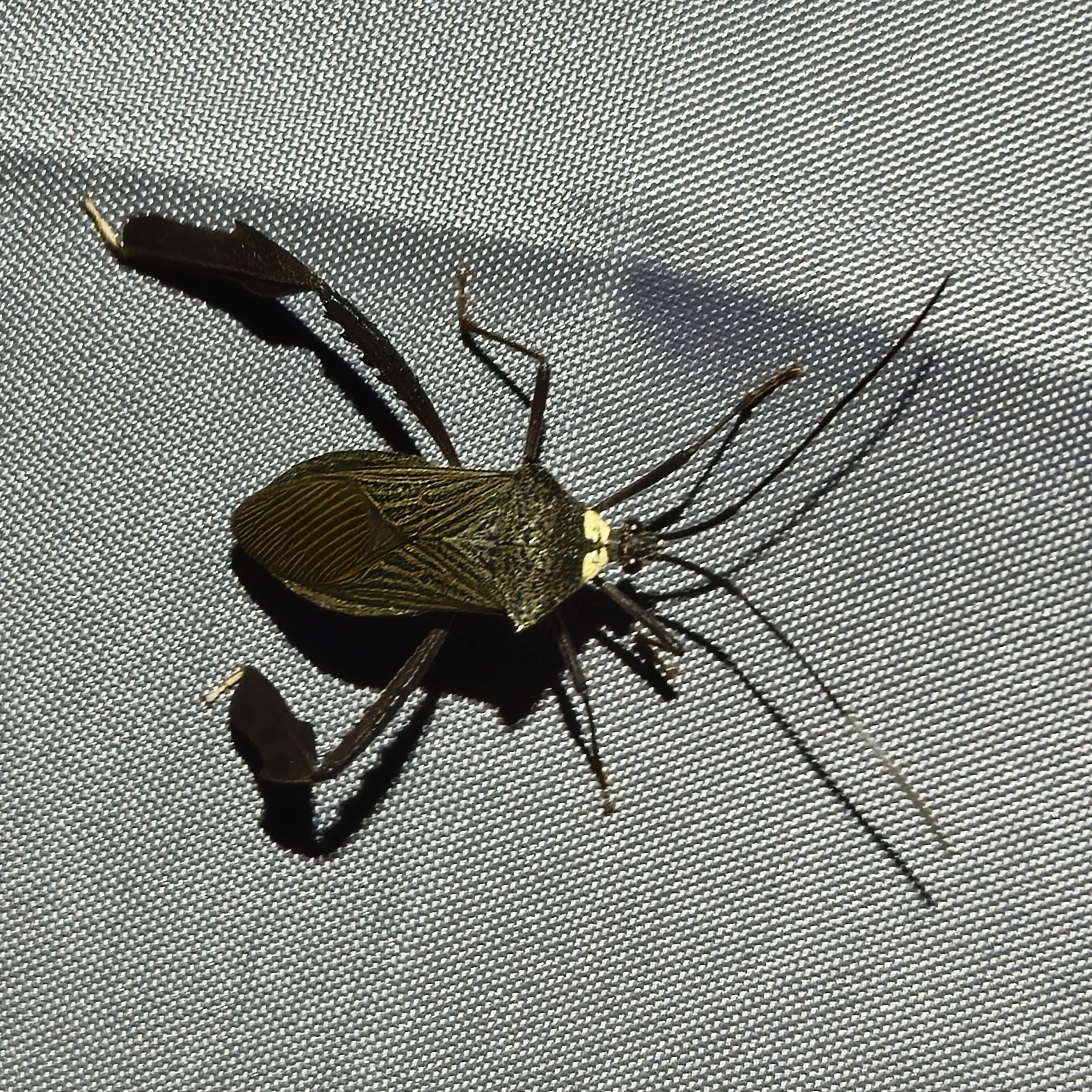}} 
& Order: \textit{Hemiptera} \textcolor{green}{\ding{51}} & Accurately identifies key hemipteran traits such as piercing-sucking mouthparts, hemelytra structure, and flattened profile. Excludes other insect orders with solid reasoning.
& Justified Classification \\
& Family: \textit{Coreidae} \textcolor{green}{\ding{51}} & Notes characteristic coreid traits like hind tibial expansions and narrow abdomen. Differentiates from similar families like Alydidae and Reduviidae.
& Justified Classification \\
& Genus: \textit{Leptoglossus} \textcolor{red}{\ding{55}} & Cites tibial leaf shape, pronotal collar, and lack of zig-zag wing pattern as justification for Leptoglossus placement.
& Misguided Justification \\
& Species: Unknown & Notes difficulty resolving species due to indistinct markings and subtle variation, leaving label as unknown.
& Abstinence \\
\hline
\end{tabular}
}
\caption{Examples of failure modes across taxonomic levels. Predictions vary in validity — some are justified, while others rely on hallucinated, non-diagnostic, or geographically implausible traits. These structured errors highlight challenges in model reasoning across biological hierarchies.}
\label{tab:failure_modes_categorized}
\end{table*}
To characterize how VLMs succeed or fail in insect classification, we conducted an expert-guided review of model predictions on $\sim$20 selected images and their accompanying explanations across a diverse set of challenging images. Specifically, we focused on cases where the expert was confident only up to a certain taxonomic level, while the model proceeded to assign finer-grained taxonomic labels. These examples (see Table~\ref{tab:failure_modes_categorized}) allow us to evaluate whether such over-specific predictions are made for biologically plausible reasons, or whether they reflect overconfidence, superficial pattern matching, or reasoning failures. In the first case, the model predicted \textit{Rivula propinqualis}. Although the expert initially considered the label plausible, closer inspection revealed that the justification relied on generic traits—such as straw-yellow coloration, resting posture, and faint forewing stripes—that are common across many unrelated genera. The expert ultimately judged the prediction unlikely, pointing out subtle inconsistencies in wing spot placement and noting that tropical members of \textit{Rivula} exhibit markedly different patterns. This case reflects clear taxonomic overreach: the model commits to a fine-grained species label based on superficial matches, without sufficient discriminative evidence. The justification also exemplifies misguided reasoning, as it depends on real but non-exclusive traits that fail to uniquely support the predicted genus.

In a second example, the model correctly identified the order as \textit{Lepidoptera}, citing scaled wings and posture, but also referenced coiled tongue and metamorphosis—traits not visible in the image—resulting in morphological hallucination. It repeated this at the family level by citing invisible tympanal organs. At the genus level, it predicted \textit{Ponometia} based on mimicry and bird-dropping resemblance—features the expert deemed unverifiable—marking a speculative inference. Finally, the species prediction (\textit{Ponometia candefacta}) was rejected in favor of \textit{Elaphria} or an undescribed species, reflecting clear taxonomic overreach.

A third example starts with a correct order-level prediction of \textit{Lepidoptera}, but the model assigns \textit{Pyralidae} based on snout-like palps and narrow forewings—generic traits found across many families, leading to misguided justification. It then predicts the genus \textit{Achroia}, which the expert found entirely mismatched in shape and pattern—an instance of severe mismatch. At the species level, the model selected \textit{Achroia grisella}, which the expert deemed implausible, reflecting taxonomic overreach. The final example shows a well-grounded prediction. The model correctly identified the order \textit{Hemiptera }and family \textit{Coreidae} using visible traits like piercing-sucking mouthparts, hemelytra structure, and hind tibial expansions—aligning with expert reasoning and reflecting justified classification. At the genus level, it predicted \textit{Leptoglossus} based on tibial shape and pronotal collar, but the expert noted inconsistencies and the absence of a diagnostic zig-zag wing pattern. This reasoning relied on generic and partially misinterpreted traits, making it an instance of misguided justification. At the species level, the model abstained due to subtle variation—an appropriate and cautious abstention.

These qualitative failures highlight how VLMs, despite producing confident and linguistically plausible explanations, often rely on non-diagnostic traits, hallucinated details, or entirely superficial patterns. Such behaviors underscore the need for evaluation frameworks that assess not just taxonomic correctness, but also the interpretability and biological validity of model reasoning. We use two prompts to evaluate model behavior under taxonomic classification tasks. The first prompt elicits only the hierarchical labels (Order, Family, Genus, Species) without any accompanying explanation, while the second prompt additionally requires the model to justify each taxonomic decision for our qualitative analysis. Both prompt variants are shown below.

\begin{tcolorbox}[title=Prompt 1: Classification Only, colback=gray!5, colframe=black!50]
\small
You are an entomologist. Your job is to classify insects into the following hierarchy:

Order: \\
Family: \\
Genus: \\
Species: \\

You must return only these four fields. If you are confident about a level, fill it in. If not, write "Unknown". Do not provide explanations, reasoning, or any other text.

Only return the output in the exact format above. No markdown, no commentary, no additional lines.
\end{tcolorbox}

\begin{tcolorbox}[title=Prompt 2: Classification with Explanation, colback=gray!5, colframe=black!50]
\small
You are an entomologist. Your job is to classify insects into the following hierarchy:

Order: \\
Family: \\
Genus: \\
Species: \\

You must return these four fields. If you are confident about a level, fill it in. If not, write "Unknown".

Once classified, separate the classification with a comma and explain why you classified at each taxonomic level in 50--60 words. Provide this explanation next to each field. Only return the output in the exact format above. No markdown, no commentary, no additional lines.
\end{tcolorbox}

\subsection{Visualizing Model Predictions Across the Taxonomic Hierarchy}
\begin{figure}[h!]
    \centering
    \includegraphics[width=\textwidth]{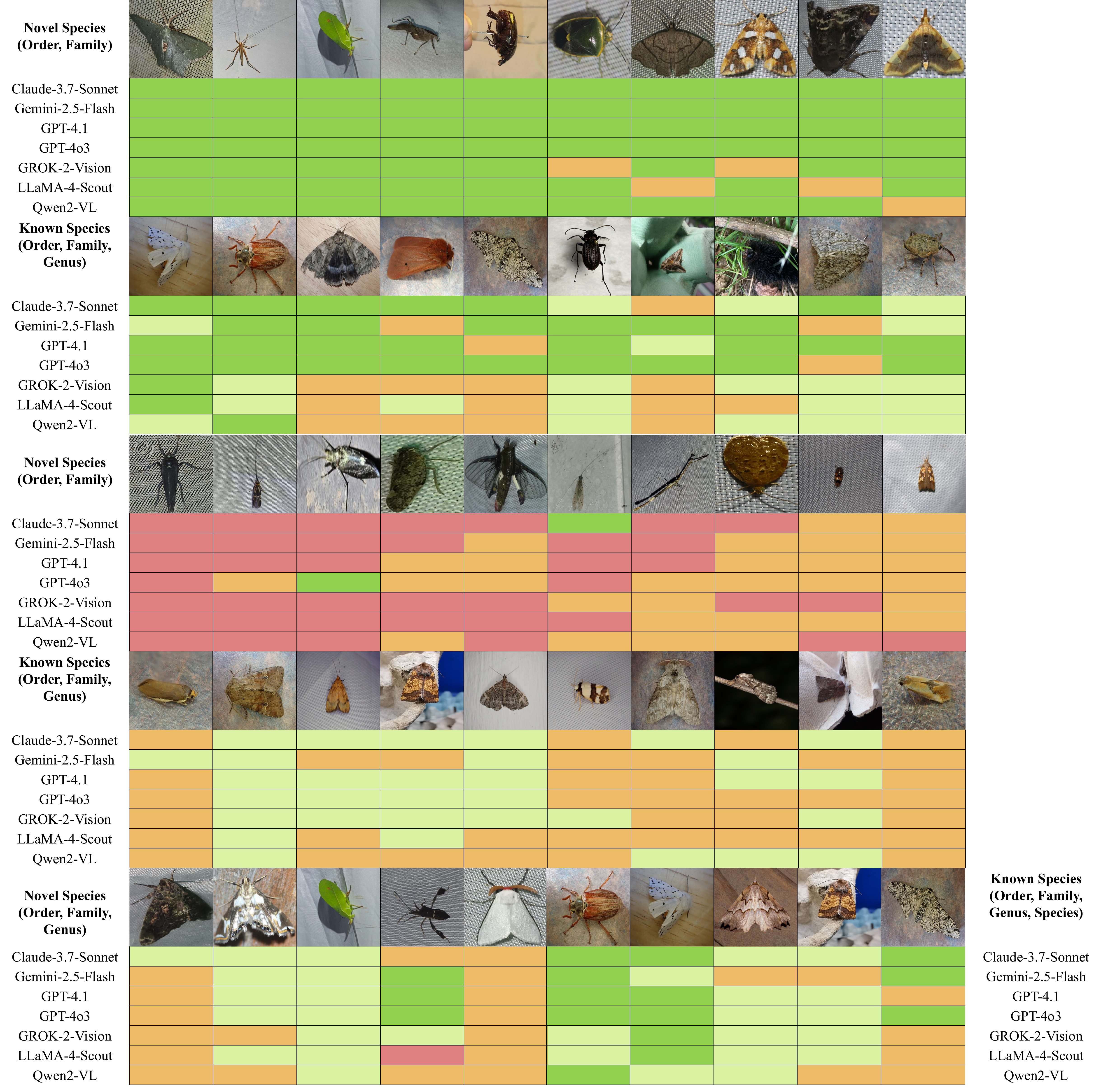}
    \caption{Taxonomic classification accuracy across models for both \textit{Known }and \textit{Novel} species.
Each panel displays per-model predictions for individual insect images, grouped by label availability—Order + Family for \textit{Novel} species (row 1, row 3), Order+Family+Genus for partially labeled novel species (row 5 \textit{(left)}), and Order + Family+Genus  (row 2, row 4) and Order + Family+ Genus+ Species for \textit{Known} species (row 5 \textit{(right)}). Rows correspond to VLMs, while columns represent individual insect images. Color-coding indicates correctness across levels: bright green (all correct), green (3 of 4 correct), orange (1–2 correct), and red (all incorrect). This visualization reveals model-specific strengths, error patterns, and generalization behavior across coarse-to-fine taxonomy.}
    \label{fig:vlm-grid-orders2}
\end{figure}

To better characterize model performance across taxonomic depth, we visualize results using color-coded heatmaps where each row corresponds to a model and each column to an insect specimen (see Figure.~\ref{fig:vlm-grid-orders2}). We evaluate predictions at four hierarchical levels—Order, Family, Genus, and Species—depending on the depth of available ground truth. We use a 4-color scheme to reflect correctness across the hierarchy. We separately analyze known and novel species due to differing label completeness:
 
\textit{\textbf{Known Species:}}All known specimens are fully labeled through the complete taxonomic hierarchy. This allows us to analyze models’ performance across all four levels. We first selected ten examples exhibiting high and low prediction accuracy at the Order, Family, and Genus levels, revealing both well-understood taxa (e.g., \textit{Lepidoptera: Noctuidae}) and frequent confusion zones (e.g., misclassifications between \textit{Erebidae} and \textit{Crambidae}).
We further highlight five examples where models attempted Species-level predictions. Performance here was predictably poor—bright green cells were rare—but some models (e.g., o3) made plausible species-level guesses on common or visually distinct taxa. The evaluation underscores the steep drop-off in performance as models traverse finer taxonomic granularity, reinforcing the challenge of species identification in zero-shot settings.

\textit{\textbf{Novel Species:}}Unlike known samples, novel species typically have partial ground truth, most commonly labeled up to Order or Family. Accordingly, our primary evaluation is at these two levels. We select ten diverse examples to probe model consistency under limited supervision. These include some high-confidence cases (e.g., consistent \textit{Coleoptera} predictions across models) and failure clusters involving rare families like \textit{Tettigoniidae}.
Additionally, we include five examples labeled up to Genus, allowing us to probe model performance slightly deeper. While some models overcommit (predicting beyond the provided taxonomy), others exhibit promising conservatism—selecting "Unknown" at lower levels despite making accurate Order/Family predictions. These examples are particularly instructive for assessing open-world calibration and uncertainty handling.

\end{document}